\pdfoutput=1
\documentclass[11pt]{article}

\usepackage[final]{ACL2023}

\usepackage{times}
\usepackage{latexsym}

\usepackage[T1]{fontenc}
\usepackage[utf8]{inputenc}

\usepackage{microtype}

\usepackage{inconsolata}
\usepackage{graphicx}

\usepackage{multicol}               % for MULTICOLUMNS
\usepackage[many]{tcolorbox}    	% for COLORED BOXES (tikz and xcolor included)

\definecolor{main}{HTML}{555555}    % setting main color to be used
\definecolor{sub}{HTML}{cde4ff}     % setting sub color to be used
\usepackage{booktabs} 

\tcbset{
    sharp corners,
    colback = white,
    before skip = 0.2cm,    % add extra space before the box
    after skip = 0.2cm      % add extra space after the box
}                           % setting global options for tcolorbox

\newtcolorbox{boxB}{
    fontupper = \color{main}, % font color
    boxrule = 1.0pt,
    colframe = main,
    rounded corners,
    arc = 5pt   % corners roundness
}

\title{Not All Personas Are Worth It: \\Culture-Reflective Persona Data Augmentation}

\author{Ji-Eun Han \\
  KT \\
Seoul, South Korea \\
  \texttt{hanji0514@gmail.com} \\\And
  Yoonseok Heo\thanks{* Corresponding author} \\
    KT \\
Seoul, South Korea \\
  \texttt{nlp.ysheo419@gmail.com} \\}

\begin{document}
\maketitle
\begin{abstract}
Incorporating personas into conversational AI models is crucial for achieving authentic and engaging interactions. However, the cultural diversity and adaptability of existing persona datasets is often overlooked, reducing their efficacy in building culturally aware AI systems. To address this issue, we propose a two-step pipeline for generating culture-specific personas and introduce KoPersona, a dataset comprising 200,000 personas designed to capture Korean cultural values, behaviors, and social nuances. A comprehensive evaluation through various metrics validates the quality of KoPersona and its relevance to Korean culture. This work not only contributes to persona-based research, but also establishes a scalable approach for creating culturally relevant personas adaptable to various languages and cultural contexts.
\end{abstract}

% \footnote{* Corresponding author}
\section{Introduction}

In everyday conversations, one’s persona naturally emerges, playing a crucial role in building connections. To mimic the human conversation, the persona plays an important role in the development of conversational AI, enabling models to engage more authentically and responsively with users. 
Besides conversational AI, incorporating persona is important in gaming, marketing, education and in various fields. 
%Not just in dialogue models, incorporating persona is important in marketing, education, and ...?

% personality 이전 연구에 대한 서술
% synthesize 하는 방법 서술: 사람 -> LLM으로 하는 방식
Interest in persona has been consistent. \citet{zhang-etal-2018-personalizing} introduces the crowd-sourced PERSONA-CHAT dataset, which consists of a variety of personas and a related chit-chat dialogue dataset that uses these personas. \citet{jandaghi-etal-2024-faithful} releases Synthetic-Persona-Chat dataset, an extension of PERSONA-CHAT dataset. They generated persona-based conversations using extended personas with LLMs. \citet{ge2024scalingsyntheticdatacreation} suggests a methodology for persona synthesis and introduces PersonaHub, a dataset of one billion synthesized personas.

\begin{figure}[t]
    \centering
    \includegraphics[width=\columnwidth]{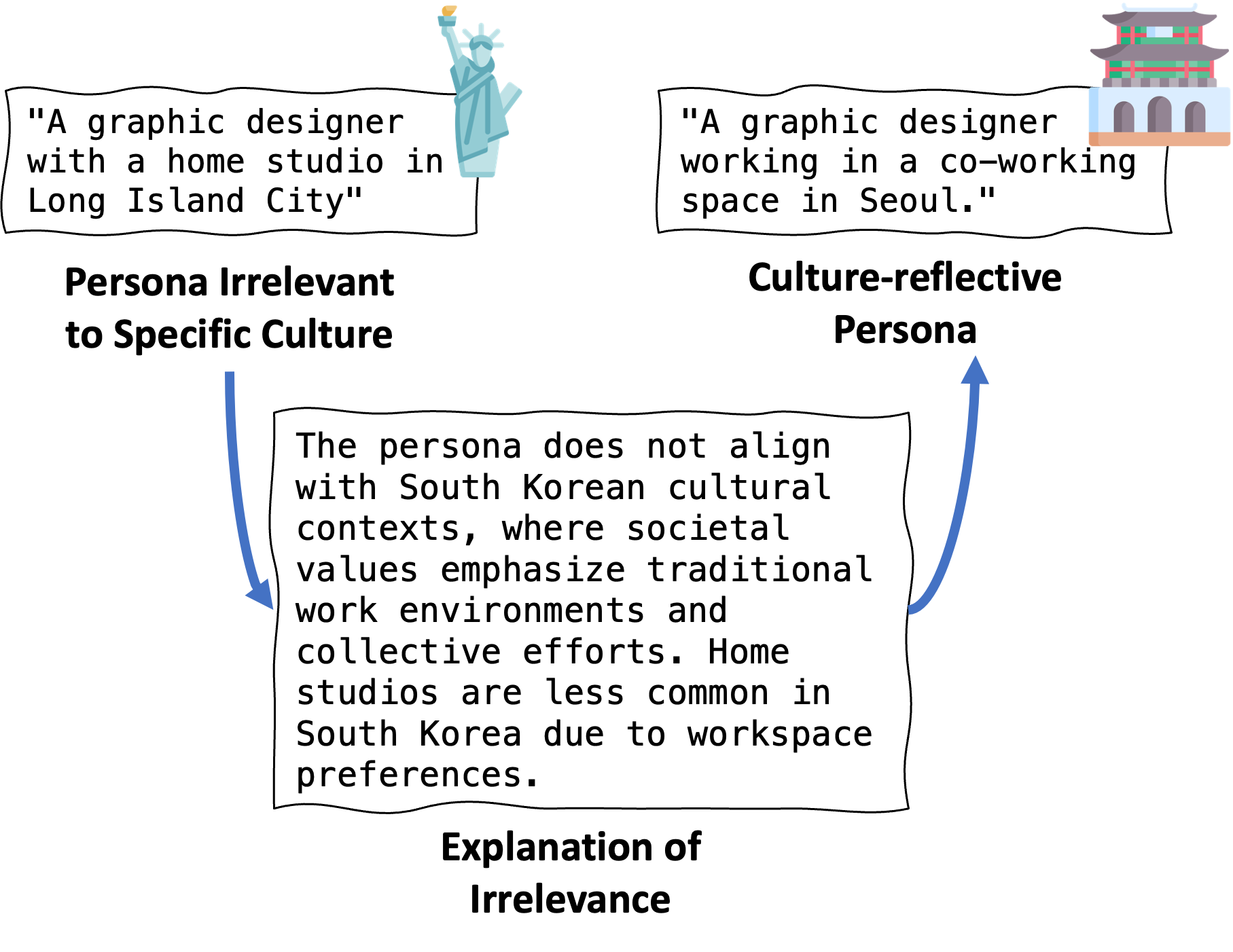}
    \caption{The process of generating culture-reflective personas by adapting a culturally irrelevant persona into a contextually appropriate one.}
    \label{fig:intro}
\end{figure}

% Motivation
Although diverse personas have been developed, their cultural applicability, essential for creating culture-specific language models, is often neglected. Figure \ref{fig:intro} illustrates the task overview of this work. For instance, we present a persona from PersonaHub: \textit{“A graphic designer with a home studio in Long Island City.”} While this persona is tailored to a U.S. cultural context, it may not be relevant or applicable to other cultures. To address this challenge, we propose a pipeline designed to generate persona datasets that align with distinct cultural settings, making them more suitable for training language-specific models. In this paper, we focus on Korean culture. By applying our pipeline, for instance, a more contextually appropriate Korean persona could be \textit{``A graphic designer working in a co-working space in Seoul.”} 
% 우리가 만든 것 설명

\begin{figure*}[ht!]
    \centering
    \includegraphics[width=\textwidth]{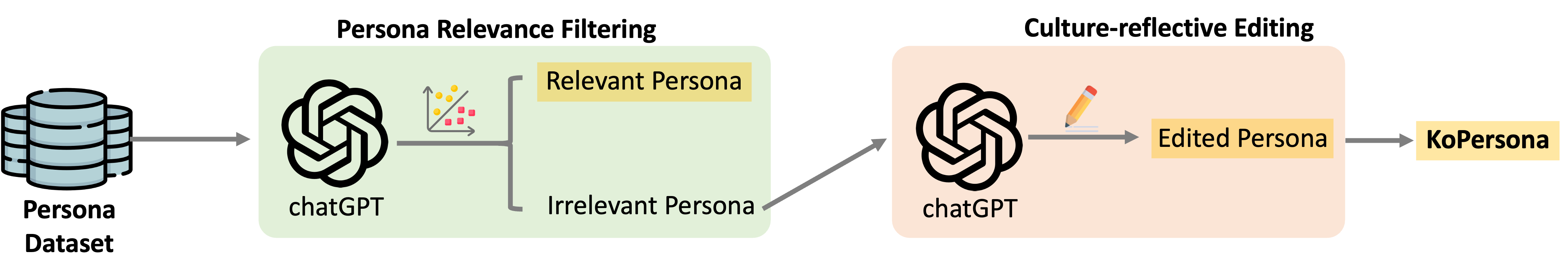}
    \caption{The overview of the suggesting pipeline}
    \label{fig:overview}
\end{figure*}
%한국은 무슨 연구가 있었는지
%한국인의 Persona 관련 연구는 없음
In the context of Korean personas, \citet{bae-etal-2022-building} constructs role-specific dialogues focused on a persona of senior care. \citet{dong2024ithemfluididentities} explores individualistic and collectivistic personas in Western and Eastern languages, showing that Eastern personas, including Korean, tend to align more closely with collectivistic values. However, neither work specifically focuses on constructing Korean-specific personas.

% Contribution
Our contributions are three-fold:
\begin{itemize}
    \item We propose a two-step pipeline to augment the existing persona dataset into culturally enriched persona, focusing on creating culture-specific personas.
    \item We release KoPersona, 200,000 personas designed to authentically reflect Korean cultural values, behaviors, and social nuances. We label the dataset into two categories: general persona and culture-reflective persona.
    \item We validate KoPersona with various metrics. The evaluation result shows that KoPersona is diverse and reflects the cultural context compared to the original dataset before applying our pipeline.
\end{itemize}

% open source 모델을 사용하여 실험을 진행하였다

% figure 인용 안됨 문제

\section{Method}
As our source data, we utilize a subset of PersonaHub introduced by \citet{ge2024scalingsyntheticdatacreation}, consisting of 200,000 personas. We selected this data set for its extensive diversity in personas. Figure \ref{fig:overview} shows the overview of our pipeline, which includes two key steps: persona relevance filtering, and culture-reflective persona editing.
Each step is designed with the specific goal of adapting personas to reflect Korean cultural values and social nuances, ensuring that the resulting dataset aligns with the linguistic and cultural context of Korea. We attempted to use open-source language models like Llama3.2 \cite{dubey2024llama3herdmodels} and Qwen2.5 \cite{qwen2.5}, but the results did not meet our expectations. Thus, we use the ChatGPT-4o mini API to implement the pipeline and achieve the desired data quality.

% augmentation pipeline
% culture reflective
% LM이 이미 korean에 대해 알고 있음

\subsection{Persona Relevance Filtering}\label{sub_personarelevance}
The initial step in the pipeline is to determine the relevance of personas from the source dataset to Korean contexts. Each persona is evaluated and classified according to its relevance to Korean circumstances. Inspired by Chain-of-Thought prompting \cite{wei2023chain}, we enhance the filtering process by having the model generate explanations for its relevance determinations for each persona.

We identified 85,878 personas as relevant and 114,122 as irrelevant. Relevant personas are applicable within the Korean context, whereas irrelevant ones frequently contain names or settings specific to other countries.

% general persona 특징
% irrelevant persona 특징
In Table \ref{tab:personas}, we categorize the characteristics of personas that are not applicable to the Korean context. It includes original personas from PersonaHub that do not correspond to the Korean context. A persona is classified as irrelevant if it includes specific locations, such as foreign countries or cities, specific cultures, nationalities, individuals, historical references, or economic conditions that are not relevant to Korea.

% Category: Represents the thematic grouping of the personas, including aspects such as location, culture, nationality, personal aspirations, historical context, and economic background.

% Persona: Describes the initial persona, often reflecting a context or perspective that may not align with South Korean cultural, societal, or economic norms.

% The first phase in the pipeline is to check the personas from the source dataset are relevant to Korean circumstances. Given a persona, we classify whether the persona is relavant to Korean circumstances. XXXXX personas are classified they are relavant and XXXXX persona are classified they are not relevant. Personas that are classified as relevant can be used universally. However, personas classified as irrelavant contain people's name or particular setting of many countries.
% TODO) 사용한 Prompt 넣기

\subsection{Culture-reflective Persona Editing}\label{sub_culture}
In the second step, we modify the personas classified as irrelevant to align them with the Korean context. This process is performed using the same model that initially determined the relevance of the persona. The model edits these irrelevant personas based on the descriptions generated during the Persona Relevance Filtering stage. 

The transformed examples are shown in Table \ref{tab:personas}. They have been adapted to better resonate with South Korean cultural values, societal dynamics, and economic realities. These adaptations ensure that the personas remain relevant and relatable within the South Korean context. For instance, locations originally set in the UK have been changed to South Korea. Cultural elements common in South Korea, such as the use of home-filtered water machines, have been incorporated. Nationalities have been updated from Polish to South Korean, and cities like Tarnowskie Góry have been replaced with Seoul. Notable figures such as Danny Boyle have been substituted with Bong Joon-ho, a South Korean film director who won an Academy Award in 2020. Historical references have been adjusted to reflect Korea's history, including its colonization by Japan. Economic backgrounds have been modified to mirror South Korean economic conditions. These changes are designed to ensure that the personas align with South Korean culture and are relevant to users within this context.

% The transformed examples are shown in \ref{tab:personas}. They are adapted to better resonate with South Korean cultural values, societal dynamics, or economic realities. The adaptations are designed to ensure the personas remain relevant and relatable to the South Korean context. Each entry demonstrates how cultural, historical, and societal nuances influence the redefinition of a persona. In terms of the location, UK has been changed into Sout Korea. In culture, home-filtered water machine is common in South Korea. Thus, this culture has been reflected. For nationality, Polish has been changed to South Korean and Tarnowskie Go/ory has been changed into Seoul. In person, Danny Boyle has been changed into Bong Joon-ho, a file director who won Academy Award in 2020. In history, Korea was colonized by Japanese so this kind of historical background has been reflected. In economy, South Korean economical background has been reflected in edited persona. 

% For instance, the edited personas incorporate locally significant figures, industries, and perspectives to enhance relevance, such as replacing global idols with South Korean counterparts or shifting economic contexts from coal mining to factory work, reflecting South Korea's modern industrial landscape.

% 왜 editing을 해야하는지

\begin{table*}[ht]
\centering
\resizebox{\textwidth}{!}{%
\scriptsize
\begin{tabular}{l|p{5cm}|p{5cm}}
\toprule
\textbf{Category} & \textbf{PersonaHub (Original Persona)} & \textbf{KoPersona (Edited Persona)} \\ \midrule
Location & A digital anthropology professor based in \textit{the UK} & A digital anthropology professor in \textit{South Korea}, \textit{exploring local traditions and global influences.} \\ \midrule
Culture & A health-conscious millennial \textit{who doesn't trust bottled water companies} & A health-conscious millennial \textit{who prefers home-filtered water over bottled options.} \\ \midrule
Nationality & A \textit{Polish} political science student living in \textit{Tarnowskie Góry} & A \textit{South Korean} political science student living in \textit{Seoul}. \\ \midrule
Person & An aspiring filmmaker deeply inspired by \textit{Danny Boyle}'s creativity and innovation & An aspiring filmmaker deeply inspired by \textit{Bong Joon-ho}'s creativity and innovation. \\ \midrule
History & A \textit{Japanese} political blogger \textit{who is personally fond of Taro Aso} & A \textit{South Korean} political blogger \textit{critical of Japan's past actions but exploring reconciliation.} \\ \midrule
Economy & A retired \textit{coal miner} who shares stories of their past experiences and offers advice on navigating the changing job market & A retired \textit{factory worker} who shares stories of their past experiences and offers advice on navigating the job market. \\ \bottomrule
\end{tabular}%
}
\caption{Original persona from PersonaHub \cite{ge2024scalingsyntheticdatacreation} and KoPersona, which is an edited version derived from PersonaHub, for enhanced cultural reflectiveness.}
\label{tab:personas}
\end{table*}

\subsection{Prompting}
We release a prompt used for creating KoPersona. The general prompt used in the pipeline is suggested as follows:

\begin{boxB}
Analyze the provided persona to determine if it aligns with South Korean cultural and social contexts. \newline \newline
PERSONA: \{{persona}\} \newline \newline
If the persona aligns, set ``Decision" to ``Yes". \newline
If the persona does not align, set ``Decision" to ``No" and provide reason why it does not align. \newline \newline
Then provide an edited version of the persona that is adapted to South Korean circumstances.
\end{boxB}
\noindent In order to reduce the cost of using the ChatGPT API, we did not execute the process in individual phases as shown in Figure \ref{fig:overview}. Instead, we provide the model with a prompt that generates all elements simultaneously.

\subsection{Construction of KoPersona Dataset}
After the Culture-reflective Persona Editing step, we combine the personas identified as relevant to Korean culture in step 1 with the edited personas from step 2. The personas from step 1 are labeled as `general personas', and the ones from step 2 are referred to as `edited personas'.

% TODO) 어떤 persona들이 irelevant로 분류되었는지 위에서 언급했는데 분석해서 쓸 필요가 있음
% TODO) 어떤 prompt를 썼는지 넣기
% Editing

% \subsection{Edited Persona Filtering}
% % 이걸 하는 이유는 뭔지
% % Editing이 어떻게 되었는지 서술
% To ensure the quality of the edited persona, we once again go through the Persona Relevance Filtering. If it is guaranteed to be relevant, then we choose this edited persona to be the final. In this way, we can guarantee the quality of the persona.

% \subsection{Persona Translation}
% We translate KoPersona into Korean, as the personas were originally constructed in English. This augmentation aims to support the training of a Korean language model.

% \subsection{Persona-based Dialogue Generation}
% We pair two random personas, assigning each as an interlocutor, and generate a multi-turn dialogue between them. For this, we use in-context one-shot dialogue generation to create the multi-turn interactions.
% We create a dialogue sample using ChatGPT-4 with a given pair of personas, which the model then uses as a reference to generate similar dialogues.
% 어떤 모델로 dialogue 를 만들었는지

% \subsection{Statistics of KoPersona}
% \begin{table}[h]
%     \centering
%     \resizebox{\columnwidth}{!}{%
%     \begin{tabular}{m{3cm} | c | c | c}
%         \toprule
%         \textbf{} & \textbf{General Persona} & \textbf{Culture Persona} & \textbf{Total} \\
%         \midrule
%         \# of personas & 85,878 & 114,122 & 200,000 \\
%         Average token length & 5.15 & 4.69 & - \\
%         \bottomrule
%     \end{tabular}%
%     }
%     \caption{Statistics of KoPersona}
%     \label{tab:statistics}
% \end{table}

We have categorized the personas into two types: general personas and culture personas. The general personas are those determined to be relevant to the Korean context during the Persona Relevance Filtering step. The culture personas, short for culture-reflective personas, are those initially irrelevant to the Korean context and subsequently modified during the Culture-Reflective Persona Editing step. Among the total of 200,000 personas, 85,878 are general personas and 114,122 are culture personas. The average token lengths for each category are 5.15 and 4.69, respectively.
% Input configuration
% Experimental detail
% Baseline model
% Evaluation metric
\section{Experiment}
We conduct both quantitative and qualitative evaluations, including comprehensive comparisons with the original persona dataset prior to applying our pipeline, to demonstrate that KoPersona is meticulously designed and significantly outperforms its predecessor in terms of cultural reflectiveness, highlighting its capacity to adapt to and accurately represent diverse cultural contexts.
\subsection{Quantitative Evaluation}
We evaluate the diversity of the original personas before applying our proposed pipeline (PersonaHub \cite{ge2024scalingsyntheticdatacreation}) and compare it to that of KoPersona. To check the accuracy of transformed personas, we train BERT \cite{devlin-etal-2019-bert} base model with 20,000 personas sampled from KoPersona. The training dataset contains 10,000 original personas and 10,000 edited personas. The model is trained to classify whether the given persona has South Korean context or not. We split train and test dataset as 7:3. We train the model for three epochs. The average testing accuracy is 0.98.

% result
\begin{table}[t]
    \centering
    \resizebox{\columnwidth}{!}{%
    \scriptsize
    \begin{tabular}{l | c |c | c}
        \toprule
        \textbf{Dataset} & \textbf{BLEU-2↓} & \textbf{Jaccard Sim↓} & \textbf{P-Acc↑} \\
        \midrule
        PersonaHub & 1.21 & 1 & 0.86 \\
        KoPersona (Ours) & \textbf{0.72} & \textbf{0.38} & \textbf{0.99} \\
        \bottomrule
    \end{tabular}%
    }
    \caption{Comparison of diversity and persona-accuracy  between PersonaHub \cite{ge2024scalingsyntheticdatacreation} and KoPersona}
    \label{tab:diversity_comparison}
\end{table}

Table \ref{tab:diversity_comparison} compares the PersonaHub dataset with our proposed KoPersona dataset in terms of diversity and cultural alignment. Diversity metrics, BLEU-2 \cite{papineni-etal-2002-bleu} and Jaccard Similarity, evaluate lexical variety and distinctiveness, where lower scores indicate greater diversity and reduced repetitiveness. Unlike standard interpretations where higher scores denote better diversity, we reverse the evaluation to consider the PersonaHub dataset as a reference. Lower scores in these metrics therefore indicate increased diversity, highlighting KoPersona’s capability to produce more varied and distinctive responses compared to the PersonaHub dataset. 
The PersonaHub dataset achieves BLEU-2 and Jaccard Similarity scores of 1.21 and 1.00, respectively, suggesting moderate diversity. In contrast, KoPersona achieves significantly lower scores of 0.72 for BLEU-2 and 0.38 for Jaccard Similarity, demonstrating improved lexical variety and uniqueness. Persona Accuracy (P-Acc) measures the dataset’s ability to reflect cultural nuances. With a P-Acc score of 0.99, KoPersona outperforms the PersonaHub dataset, which achieves a lower score of 0.86. These results highlight that KoPersona is better suited for applications requiring culturally adaptive responses, as it balances diversity with cultural alignment.
\subsection{Qualitative Evaluation}
As a qualitative evaluation, we evaluate 500 personas from PersonaHub \citep{ge2024scalingsyntheticdatacreation} and KoPersona using similar approach to G-Eval \cite{liu2023gevalnlgevaluationusing}. The prompt we use for the evaluation is suggested as follows:
\begin{boxB}
Based on the provided persona, rate how well this persona aligns with the South Korean cultural context on a scale of 1 to 5. Justify your rating with explanation. \newline \newline
Persona: {persona}
\end{boxB}

\begin{table}[h!]
\centering
\resizebox{\columnwidth}{!}{%
\scriptsize
\begin{tabular}{l|c}
\toprule
\textbf{Persona} & \textbf{ Average Rating} \\ \midrule
PersonaHub \cite{ge2024scalingsyntheticdatacreation}  & 3.05 \\  
KoPersona (Ours)  & \textbf{4.55} \\ \bottomrule
\end{tabular}
}
\caption{Ratings for Personas with Average Rating}
\label{tab:ratings_with_average}
\end{table}
\noindent Table \ref{tab:ratings_with_average} presents a comparison between the PersonaHub dataset and KoPersona regarding their cultural alignment with Korea, as evaluated by ChatGPT. The average rating for PersonaHub is 3.05, while KoPersona received a higher average score of 4.55. This indicates that KoPersona aligns more closely with Korean culture.

% The table \ref{tab:ratings_with_average} compares two persona dataset, 1 Billion and KoPersona in terms of the chatGPT evaluation of aligned with Korean culture. 1 Billion" has been rated with an average score of 3.05, while KoPersona has a higher average rating of 4.55. The higher rating for ours suggests that KoPersona is viewed more favorably or aligns better with the Korean context.
% culture reflective 어떻게 보일 것인가
% -> Embedding Analysis for Measuring Cultural Similarity
% Kopersona로 train한 classifier

% 비교 대상
% 기존 persona | edited persona (diversity 비교)
% dist 1, 2 로 비교
% 만든 dialogue가 Kopersona를 가지고 있는지 아닌지 비교
% kopersona로 train한 cls가 판단
%https://github.com/PaddlePaddle/models/blob/release/1.6/PaddleNLP/Research/Dialogue-PLATO/plato/metrics/metrics.py

% 만든 dialogue가 kopersona 가지고 있는지, fluency, 등등
% G-eval

\section{Conclusion}

% We present a framework for creating culture-reflective persona dataset and dialogues. 
% 5 step pipeline

% We concentrate solely on creating personas aligned with Korean culture. However, our framework is adaptable, enabling users to create personas specific to their own cultural context with ease.

% % Limitation
% Rely on the one large language model to edit the persona. maybe biased

In this paper, we propose a pipeline for augmenting personas to create culture-specific datasets, focusing on adapting existing personas to align with the Korean context. We have addressed the need for culturally specific persona datasets by introducing KoPersona. It is a collection of 200,000 personas designed to reflect Korean context. Our approach involves two pharse - filtering irrelevant personas and editing them to reflect Korean culture. We validate the dataset through various metrics to ensure its quality and relevance to Korean culture.

KoPersona fills a significant gap in resources available for Korean-specific language model training, enabling the creation of models that are more culturally sensitive. By incorporating locally significant figures, industries, and perspectives, we have ensured that the personas are relevant to Korean circumstances. Our work not only introduces a valuable dataset to the research community but also establishes a scalable framework for generating culture-specific personas across diverse linguistic and cultural contexts. By extending this framework to accommodate diverse cultural elements and adapt to various contexts can further enrich the diversity and scalability of models.

\bibliography{custom}
\bibliographystyle{acl_natbib}

% \appendix

% \section{Example Appendix}
% \label{sec:appendix}

\end{document}